\documentclass[sigconf]{acmart}
\AtBeginDocument{%
  \providecommand\BibTeX{{%
    \normalfont B\kern-0.5em{\scshape i\kern-0.25em b}\kern-0.8em\TeX}}}

\setcopyright{acmcopyright}
\copyrightyear{2022}
\acmYear{2022}
\acmDOI{XXXXXXX.XXXXXXX}

\acmBooktitle{HRI 2022 Workshop: Robo-Identity: Exploring Artificial Identity and Emotion via Speech Interactions, March 7-10, 2022, Sapporo, Hokkaido, Japan}
\acmPrice{15.00}
\acmISBN{978-1-4503-XXXX-X/18/06}

\begin{document}

\title{Robotic Speech Synthesis: Perspectives on Interactions, Scenarios, and Ethics}

\author{Yuanchao Li}
\email{y.li-385@sms.ed.ac.uk}
\affiliation{%
  \institution{School of Informatics, University of Edinburgh}
  \country{UK}
}

\author{Catherine Lai}
\email{c.lai@ed.ac.uk}
\affiliation{%
  \institution{School of Philosophy, Psychology \& Language Sciences, University of Edinburgh}
  \country{UK}
}

\renewcommand{\shortauthors}{Li, et al.}

\begin{abstract}
In recent years, many works have investigated the feasibility of conversational robots for performing  specific tasks, such as healthcare and interview. Along with this development comes a practical issue: \emph{how should we synthesize robotic voices to meet the needs of different situations?} In this paper, we discuss this issue from three perspectives: 1) the difficulties of synthesizing non-verbal and interaction-oriented speech signals, particularly backchannels; 2) the scenario classification for robotic voice synthesis; 3) the ethical issues regarding the design of robot voice for its emotion and identity. We present the findings of relevant literature and our prior work, trying to bring the attention of human-robot interaction researchers to design better conversational robots in the future.
\end{abstract}




\maketitle

\section{Introduction}
The rapid development of speech synthesis in recent years has made it possible for computers to generate speech that closely resembles human speech, including emotional voice \cite{lee2017emotional,tits2019exploring}. Robotics is one of the applications that benefit most from this technology. Since the facial expressions and body movements of robots are currently difficult to make as natural as those of human beings, changes in the voice are the most common way to adapt to different scenarios \cite{walters2008human,bedaf2018multi}. In this paper, we first illustrate the importance of prosody to emotional speech synthesis and the difficulty of synthesizing interactional aspects of speech. Through our prior work, we discuss the prosodic and emotional aspects of backchannels, as it is an important component of spoken dialogue that have been widely studied in human-robot conversation \cite{hussain2019speech, al2009generating, park2017backchannel}. Next, we present our designed scenario classification for robotic speech synthesis and how people perceive robot voices in different scenarios. Finally, we discuss some ethical issues, focusing on the question that how should we design robotic voices.

\section{Interactional Speech Synthesis}
\subsection{Prosody Settings for Emotion and Identity}
Prosody has been proven a dominant aspect to convey emotions.
% Among all the prosodic features, pitch, timing, and loudness have been shown to correlate with the expression of emotion through vocal prosody \cite{vinciarelli2008social}.
And in fact, prior work has succeeded in setting prosodic features for emotional speech synthesis. In \cite{burkhardt2000verification}, F0 mean (F0: fundamental frequency on which human perception of pitch depends), F0 range and tempo were increased by 50\%, 100\% and 30\% respectively for expressing joy, and 150\%, 20\% and 30\% for fear in German. The synthesized emotions were successfully judged by native speakers. \cite{murray1995implementation} found that increasing F0 mean, F0 range, tempo, and loudness by 10 Hz, 9 semitones, 30 words per minute, and 6 decibels (resp.) is suitable for expressing anger in British English. \cite{bone2014robust} found log-pitch, intensity, and loudness have high correlations with arousal in four emotion corpora. In our prior work, we also observed valence has clearer correlation with pitch-related features (low, narrow and wide pitch) than other prosodic features, and arousal is also correlated with intensity and speaking rate in addition to pitch \cite{li2017emotion}.

Neural text-to-speech has been widely used \cite{ling2015deep}, yet adjusting prosody settings using Speech Synthesis Markup Language (SSML) \cite{taylor1997ssml,baggia2010speech,walker2001new} to specify parameters for intended emotion is still a typical way for robots, considering the difficulties of real-world applications. Similar to emotion, robot identity can also be formed by prosody variation. Robot identity is a long-term signal usually contains gender, age, personality, etc., and they can also be included in the SSML settings \cite{bendel2017ssml}.

However, the majority of prior work focused on synthesizing monologues but ignored the aspects of spoken interaction which is necessary for emotion and identity. Personality is largely dependent on fillers and backchannels \cite{bruck2011impact,yamamoto2018dialogue}, whose pragmatic meanings are highly conveyed by prosody \cite{ward2016interactional}. Taking the backchannel ``really'' as an example, it can be used for expressing interest, surprise, or disappointment, depending on its speaking style. In conversations, utterance amount, backchannel frequency, filler frequency, and switching pause length have proven relevant to robot personality traits \cite{yamamoto2018dialogue}. \cite{ohshima2015conversational} revealed that subjects who were less socially adept reported feeling that their robot interlocutor was more sincere than its human counterpart because the robot’s conversational fillers helped mitigate awkwardness and express a cooperative attitude during the interaction.
% Therefore, synthesizing this sort of interactional speech remains difficult from a practical perspective.
Here, we take backchannels as an example for discussion as we have studied it in human-robot conversation.

\subsection{Backchannels}
In a prior work \cite{li2019expressing}, we used a female humanoid to conduct spontaneous chats with student participants. The humanoid was remotely operated by a female human operator in a Wizard of Oz manner. The conversations between the humanoid and participants were recorded and analyzed. We focused on the following backchannels: ``Really?'', ``Ah, I see.'' and ``I get it.'' because we found they occur most frequently in the conversations. Our subjective evaluation demonstrated that when generated in emotions mimicking the participants, the robot backchannels were perceived as natural and authentic compared to those without emotions. What's more, we also found that even when the backchannels were generated with random emotions, the participants still felt the robot feedback more natural than that without emotions.

The difficulties in synthesizing backchannels are not only the lexical form, but also prosody and timing. False prosody or timing may cause a fatal problem in HRI. For example, imagine if a robot responds “Really?” in a happy voice when the user in fact feels sad and seeks sympathy, then the user may not want to talk to the robot anymore. The solutions lie in 1) accurately recognizing the user's emotion and matching it, 2) accurately recognizing when to respond. Regarding 1), we noticed that even a simple emotion mapping using only happiness and disappointment corresponding to high and low valence can achieve satisfactory user experience \cite{li2017emotion}. Regarding 2), some predictive rules have been established. For example, for English backchannel generation is: Upon detection of P1. a region of pitch less than the 26th-percentile pitch level and P2. continuing for at least 110 milliseconds, P3. coming after at least 700 milliseconds of speech, P4. providing you have no output backchannel feedback within the preceding 800 milliseconds, P5. after 700 milliseconds wait, backchannels should be produced. For Japanese, some parameters are different: P1 = 28, P2 = 110, P3 = 700, P4 = 1000, and P5 = 350 \cite{ward2000prosodic}. As technology advances, however, these solutions will change as well.

\subsection{Open Challenges}
While some companies, such as Google, have successfully used synthetic speech to ``fool'' humans on the phone\footnote{https://ai.googleblog.com/2018/05/duplex-ai-system-for-natural-conversation.html}, it is still too early to say that robotic speech synthesis has matured to the point where it can respond naturally. We list the following challenges:

1. For utterances with clear lexical meanings, occasional prosody errors may not have a large effect on one's evaluation of the robot. However, people have a much lower tolerance for the errors when it comes to interactional speech whose interpretation largely depends on prosody, such as backchannels.

2. Detection of positions for feedback utterances is more difficult than that of other types of dialogue turns. For example, backchannel generation requires a system to be able to process lexical and non-lexical aspects of an utterance, and sometimes involves related tasks such as turn-taking detection.
% should be generated at the short pause of the speaker while not taking his/her turn, which involves related detection tasks such as turn-taking detection.

3. Compared to synthesizing isolated sentence utterances (the commonly used evaluation criteria for text-to-speech), it's hard to develop a uniform rule for evaluating synthesized feedback utterances as they are highly context dependent.
% Some people prefer fast and accurate responses from robots, while some prefer human-like behaviors even they contain errors.

4. Different languages and cultures have different conventions for conversing. Some languages, such as Japanese, have a large number of fillers and backchannels. How cultural differences should be handled also needs to be taken into account when synthesizing this type of speech.

\section{Scenario Classification}
To the best of our knowledge, there are no uniform standards or best practices for designing robotic voices, not to mention its identity, and the majority of results can only represent scenarios set by the experiments themselves. Here, we present our scenario classification in robotic voice evaluation and explore the differences in their respective focus and the challenges that exist.

\subsection{General Scenario}
The first category is the general scenario, which typically investigates robot voice in spontaneous dialogues without considering specific applications. For example, in our past work, we recorded and analyzed daily conversations between human participants, whose topics include greetings, introductions, hobbies, daily life, and a little impromptu banter. Based on the analysis results, we designed the robot's emotional voice by changing the prosody and had human participants engage in conversations with it on the same topics. The evaluation results showed that the human participants were satisfied with the robotic voice and found it to be very emotionally realistic \cite{li2019expressing}. We believe such robots can be used for all kinds of daily chats without considering specific applications.

When designing this type of robot voice, the most critical point is that the voice should be real as in human sounding and meet the user's preference, so that the user can feel comfortable. Past studies support this viewpoint from some aspects: Users prefer robot voice whose gender (male/female) matches their own \cite{eyssel2012if,lee2000can}, and are more attracted by the robot having similar personality traits (introversion/extroversion) \cite{nass2005wired}.

\subsection{Application-Dependent Scenario}
We define the second category as the application-dependent scenario, which focuses on designing robotic voices for particular uses and expressing its identity firmly. In such cases, the emotions that can be expressed are limited (e.g., sad emotions that can cause discouragement may not be allowed in some applications such as team sports \cite{lala2017utterance}). Taking healthcare scenario as an example, \cite{james2020empathetic} synthesized a flat monotone and an empathetic voice via prosodic variation for a robot named Healthbot. They recruited 120 participants and asked them to share their perceptions after watching videos of Healthbot talking in the two voices. The results reflected that people prefer empathetic voice for healthcare robots.

Another example is the instruction scenario. \cite{edwards2019evaluations} recruited student participants to evaluate a new robotic operating system by rating their perceptions towards the robotic voice, which was created as older male voice by using a text-to-speech program. Ten participants were asked to indicate the perceived age of the robotic voice. Results indicated that higher age-identified students rated the older robotic voice higher for credibility, social presence and reported more motivation to learn. Perhaps the older male voice sounds like a “professor” or “instructor” role identity.

We believe these phenomena are caused by consensus, where we unconsciously assume that certain voices and languages have specific identity/professional attributes \cite{eckert2001style,hansen1997social,eastman1985establishing}.
% Thus, application scenarios need to be considered when designing robotic voices.

\subsection{Culture-Dependent Scenario}
Many past studies have demonstrated differences in the perception of robots by participants from different countries \cite{lim2021social,haring2014perception,kaplan2004afraid}. Nevertheless, few studies have discussed in depth the reasons behind this, leading to an open question: how to design robotic voices for different regional/cultural populations. Accordingly, we define the third category as the culture-dependent scenario.

Take Japan, for example, which is the most popular country in the world for robots. Japan has been influenced by Confucianism, Buddhism, and indigenous Shintoism for a long time, and believes that every object has a soul. Based on this culture, Japanese robots such as ASIMO \cite{shigemi2018asimo} and ERICA \cite{glas2016erica}, are generally more like real people. Besides, as the birthplace of manga comic culture, Japanese are very accepting of non-human features. For instance, the voice of humanoid Pepper in SoftBank stores is designed to resemble an anime robot. Moreover, according to the Global Gender Gap Report\footnote{http://reports.weforum.org/global-gender-gap-report-2020/the-global-gender-gap-index-2020/results-and-analysis/}, Japan has the largest gender gap among all developed countries, and this is reflected in the labor force. The ratio of male to female labor force is 73\%, and the vast majority of front desk, reception, and wait staff are female. Perhaps due to this special culture, receptionist robots in Japan are designed to have a female voice almost by default.

Therefore, the evaluation results obtained by participants from one cultural background cannot simply be applied to other situations.
% which should be considered when designing robotic voice.

\subsection{Open Challenges}
Unlike interactional speech synthesis, which focuses on technical challenges, scenario classification considers more design-level challenges:

1. The perception of robot emotion and identity is not only affected by its voice but also visual appearance. Giving a mismatched voice to a robot might introduce a confounding effect \cite{mcginn2019can}.

% However, mass-produced robots have the same appearance, such as Pepper, but are used in different applications. How can different voices be synthesized for the same appearance to make the user feel that the robot's emotion and identity realistic?

2. Is there any other classification design that can more comprehensively cover the scenario of robotic speech synthesis?

3. Associating voice with identity perception will inevitably raise ethical and social debates. Does assigning a robot voice based on human experience create bias?

\section{Ethical and Social Issues}
People are not yet too concerned about current problems that exist in robotic speech and have a high tolerance for robot speech errors. Based on developments in the field of speech technology and HRI, we list the following ethical issues that we think will be gradually discussed in near future.

\subsection{Language Bias}
It is well known that robotic speech synthesis usually follows automatic speech recognition. Therefore, one of the major issues of robotic speech synthesis is caused and shared by the language bias of speech recognition. Speech recognition has been developed for many years yet is still limited by the availability of training data. Languages that spoken by a large population, such as English and Chinese, are relatively mature for speech recognition and can achieve similar performance as human speech recognition in specific scenarios. However, the performance for minority languages remains high, which inevitably leads to inaccurate robotic speech synthesis and thus affects the user experience. The same is true for accents and dialects.

\subsection{Identity Bias}
As mentioned in Section 3.2, older male voice are considered to have higher credibility and more suitable for instructor robots in some cases. Similar situation also occurs with other robotic identities, such as caregivers. Is this practice of tying voice to identity harmful for society? If these settings are widely used, will they in turn affect career choices in human society?

\subsection{Gender Bias}
As mentioned in Section 3.3, there are serious gender differences in some regions, resulting in the robotic voice that uses almost exclusively one gender in certain situations. Even if this is culturally acceptable to the majority of the local population, is there a gender bias involved in such design? Conversely, if the voice of male and female were used fairly, would the people of the region be willing to accept these robots?

\subsection{Aesthetic Bias}
In today's society, where aesthetic diversity is prized, will the aesthetics and preferences of voice also become a topic of discussion? If so, how should we design voice timbre when synthesizing robotic voices? Should they be associated with the appearance of the robot? For example, should a robot with a cute appearance always generate a sweet tone?

\section{Conclusion}
In this paper, we present new perspectives to answer this question: \emph{how should we synthesize robotic voices to meet the needs of different situations?} Firstly, we point out the difficulties of synthesizing interactional utterances that are highly frequent in dialogue speech, particularly backchannels. Secondly, we provide a novel scenario classification scheme for speech and robot researchers to better design robotic voices. Lastly, we discuss some ethical and social issues that may have not been mentioned yet. We hope our discussion can bring attention in HRI community for better robotic speech synthesis.

\begin{acks}
The first author would like to thank Kawahara Lab (Kyoto Univ.), ATR Ishiguro Lab, and Prof. Nigel Ward for their guidance in the completion of the prior work.
\end{acks}

% \section*{Short Bios}

% \, \, \, \textbf{Yuanchao Li} is a PhD student at the University of Edinburgh, fully funded by the School of Informatics. He is working on applying spoken language processing and multimodal machine learning to affective computing, human-robot interaction, and digital health.

\bibliographystyle{ACM-Reference-Format}
\balance
\bibliography{sample-base}

\end{document}